\documentclass[acmtog]{acmart}
\acmSubmissionID{1064}

\usepackage{multirow}
\usepackage{dblfloatfix} 
\usepackage{makecell}
\usepackage[caption=false]{subfig}
\usepackage[normalem]{ulem}
\usepackage[ruled]{algorithm2e} 

\usepackage{pgfplots} 
\usepgfplotslibrary{dateplot} 
\pgfplotsset{compat=newest}
\definecolor{ModernBlue}{RGB}{0, 114, 189}
\definecolor{ModernOrange}{RGB}{217, 83, 25}
\SetAlFnt{\small}
\SetAlCapFnt{\small}
\SetAlCapNameFnt{\small}
\SetAlCapHSkip{0pt}

\acmConference[]{SIGGRAPH Conference Papers ’24}{July 27–August 01, 2024}{Denver, CO, USA}

\acmYear{2024}
\acmISBN{}
\acmDOI{}

\setcopyright{rightsretained}
\setcctype[4.0]{by}

\begin{document}
\title{Radar Fields: \\ Frequency-Space Neural Scene Representations for FMCW Radar}

\author{David Borts}
\email{dborts@princeton.edu}

\author{Erich Liang}
\email{erliang@princeton.edu}
\affiliation{%
  \institution{Princeton University}
  \country{USA}
}

\author{Tim Br\"odermann}
\email{tim.broedermann@vision.ee.ethz.ch}
\affiliation{%
  \institution{ETH Z\"urich}
  \country{Switzerland}
  }
\author{Andrea Ramazzina}
\email{sefanie.walz@mercedes-benz.com}
\author{Stefanie Walz}
\email{andrea.ramazzina@mercedes-benz.com}
\affiliation{%
  \institution{Mercedes Benz}
  \country{Germany}
  }
\author{Edoardo Palladin}
\email{edoardo.palladin@torc.ai}
\affiliation{%
  \institution{Torc Robotics}
  \country{Germany}
}
\author{Jipeng Sun}
\email{js2694@princeton.edu}
\affiliation{%
  \institution{Princeton University}
  \country{USA}
  }

\author{David Bruggemann}
\email{brdavid@vision.ee.ethz.ch}

\author{Christos Sakaridis}
\email{csakarid@vision.ee.ethz.ch}
\author{Luc Van Gool}
\affiliation{%
  \institution{ETH Z\"urich}
  \streetaddress{Sternwartstrasse 7}
  \city{Z\"urich}
  \country{Switzerland}
  \postcode{8092}
  }
\email{vangool@vision.ee.ethz.ch}

\author{Mario Bijelic}
\email{mario.bijelic@princeton.edu}
\author{Felix Heide}
\email{fheide@princeton.edu}
\affiliation{%
  \institution{Princeton University}
  \streetaddress{35 Olden St}
  \city{Princeton}
  \state{New Jersey}
  \country{USA}
  \postcode{08540}
  }
\affiliation{%
  \institution{Torc Robotics}
  \city{New York City}
  \state{New York}
  \country{USA}
  }

\begin{teaserfigure}
\centering
\includegraphics[width=\textwidth]{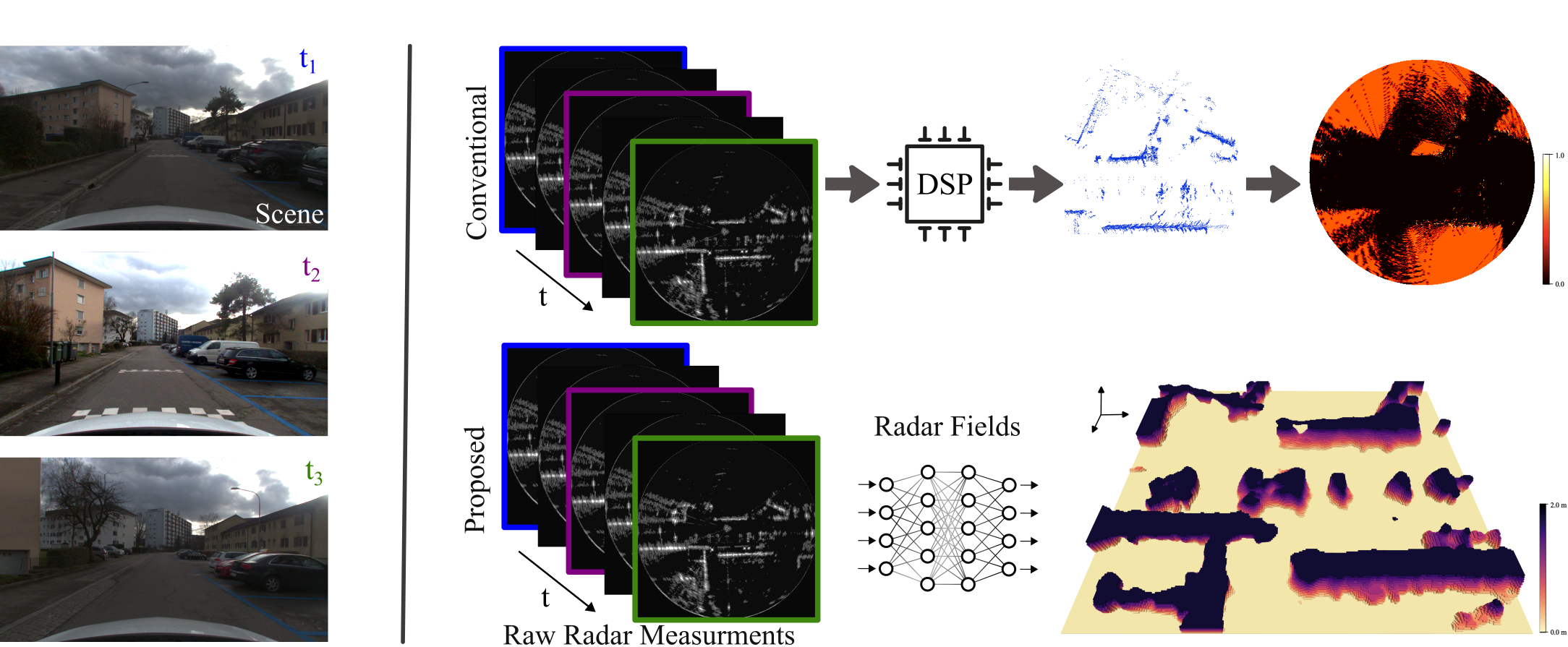}
    \caption{Conventional Radar Processing and Radar Fields. Consider a typical urban scene with complex geometry (left), including vehicles and static structures. Existing radar methods threshold raw radar FFT measurements using a DSP pipeline (top row), producing a sparse 2D bird's eye view point cloud that is fed into grid mapping algorithms to recover drivable space. Radar Fields is a neural scene representation method for Frequency-Modulated Continuous-Wave (FMCW) radar. Operating on raw frequency-domain measurements, Radar Fields can recover dense 3D occupancy from 2D radar scans (bottom row).}
    \label{fig:radar-lidar-resolution}
\end{teaserfigure}

\begin{abstract}
Neural fields have been broadly investigated as scene representations for the reproduction and novel generation of diverse outdoor scenes, including those autonomous vehicles and robots must handle. While successful approaches for RGB and LiDAR data exist, neural reconstruction methods for radar as a sensing modality have been largely unexplored. Operating at millimeter wavelengths, radar sensors are robust to scattering in fog and rain, and, as such, offer a complementary modality to active and passive optical sensing techniques. Moreover, existing radar sensors are highly cost-effective and deployed broadly in robots and vehicles that operate outdoors.
We introduce Radar Fields –- a neural scene reconstruction method designed for active radar imagers. Our approach unites an explicit, physics-informed sensor model with an implicit neural geometry and reflectance model to directly synthesize raw radar measurements and extract scene occupancy. The proposed method \textit{does not rely on volume rendering}. Instead, we learn fields in Fourier frequency space, supervised with raw radar data. We validate the effectiveness of the method across diverse outdoor scenarios, including urban scenes with dense vehicles and infrastructure, and in harsh weather scenarios, where mm-wavelength sensing is especially favorable.

\end{abstract}

\ccsdesc[300]{Computing methodologies~3D imaging}
\ccsdesc[500]{Computing methodologies~Vision for robotics}
\ccsdesc[500]{Computing methodologies~Computational photography}

\settopmatter{printacmref=false, printccs=false, printfolios=false}
\maketitle

\section{Introduction}
\label{sec:intro}
Accurate reconstruction of large-scale outdoor scenes is essential for the development and validation of self-driving robots and drones. Outdoor reconstructions lay the foundation for in-depth scene understanding, reliable navigation, and meticulous dataset generation and validation. While the large body of existing scene reconstruction methods relies on RGB and LiDAR sensors~\cite{wang2023fegr,guo2023streetsurf,huang2023nfl,tao2023lidarnerf,zhang2023nerflidar}, the potential of radar in this arena remains largely untapped. Typical radar sensors operate at around 77GHz, which offers a sensing modality complementary in two ways: an exceptional capability to detect metallic objects and a resilience against adverse weather~\cite{Bijelic_2020_STF,CramNet}, e.g., for snow or fog particle sizes which fall below the mm-wavelength. This makes radar sensing essential in urban settings cluttered with vehicles and infrastructural elements, and in scenarios compromised by rain, fog, or snow, where optical modalities can falter.

These capabilities have been instrumental in the adoption of radar sensors across multiple fields, including automotive \cite{RadarInAutonmousCars,nuscenes}, aerospace \cite{RadarAerospace}, robotics \cite{RadarRobotics}, non-invasive medical imaging \cite{RadarSiggraph2}, and human-machine interfaces \cite{RadarGestureRecognition, RadarSiggraph1}.
As an active sensor at lower cost than LiDAR -- an order of magnitude for automotive systems --  radar has emerged as a cornerstone for safety and efficiency in driver-assistance systems \cite{nuscenes,RadarInAutonmousCars}, with ever-growing adoption rates \cite{McKinsey,StatistaAutonmousSensors} predicted to match 30 million units sold by 2030.  Similarly, for robotics, the reliability and accuracy of radar data make it a vital sensor modality, both in indoor \cite{RadarRobotics, RadarIndoor1, RadarIndoor2} and outdoor \cite{RadarOutdoor1, RadarOutdoor2} settings.

Recent advances in neural rendering and scene reconstruction, epitomized by neural rendering methods \cite{mildenhall2020nerf,wang2021neus}, have relied primarily on RGB camera images. However, the performance of these methods is only as good as the quality of the input sensor data, essentially inheriting the fundamental limitations of this imaging technique. As such, existing neural rendering methods can fail in challenging light conditions  \cite{mildenhall2022nerfdark,wang2023lightingnerf}, or in the presence of challenging environmental factors, like fog \cite{levy2023seathru,ramazzina2023scatternerf}, as cameras fail to accurately capture the scene. Radar data, due to its inherent resiliency, offers a promising avenue for more versatile approaches to scene reconstruction, ensuring accuracy and reliability even in harsh weather conditions.

The potential of radar in 3D scene reconstruction is largely unexplored. This is at least partly because recovering dense geometry from radar scans is not a straightforward task. Due to the relatively large wavelengths used -- three orders of magnitude larger than LiDAR wavelengths -- FMCW radar angular resolution is lower and point targets are drastically sparser than with other sensor modalities, like LiDAR or RGB.
Moreover, many common radar sensors are just 2-dimensional, providing only azimuth-resolution with limited elevation information. These types of radars do not produce 3D point clouds.
Rather, as the beam diverges, signals from objects are mixed into a single return at each range.
This leads to a flattened 2D scan of the environment and makes it challenging to recover a 3D scene representation without a radar-specific signal formation model.
Radar data is also uniquely challenging in its sensitivity to reflective objects, as direction-dependent sensor saturation artifacts are difficult to disentangle from scene occupancy.
Altogether, these challenges render existing methods and their forward models ineffective for reconstructing scenes from radar data.

In this work, we propose a neural reconstruction method for raw radar data.
Our approach bypasses both the resolution limitations of processed radar data and the computational cost of volume rendering by modeling in \textit{frequency space}. 
We supervise our model with raw radar waveform data and recover relationships between detected power, scene geometry, distance from the sensor and scene reflectivity.
This allows our method to access the high radial resolution of the correlated waveform, which is otherwise unavailable to existing methods that operate on post-processed range estimates.

To learn in frequency space, we introduce a FMCW radar signal formation model, which draws from the physics of these sensors to differentiably decompose the received radar power at a specific distance into occupancy and reflectance-dependent components that can be learned and predicted by neural networks.
With geometry disentangled from view and material-dependent reflectance, our model can reconstruct dense scene geometry in addition to synthesizing radar waveforms, which would otherwise be impossible without such a decomposition.
In order to leverage our sensor model effectively, we train on a new dataset of raw frequency-space radar measurements, allowing us to directly predict received power at specific ranges, instead of simply predicting depth-per-ray.

Specifically, we make the following contributions in this work:
\begin{itemize}
    \item We present a neural reconstruction method for active radar sensors, which both recovers scene geometry from a sequence of radar scans and leverages its implicit scene model to synthesize raw radar data from novel views.
    \item We devise an optimization method which fits the model directly to raw frequency-space radar measurements without requiring volume rendering.
    \item We introduce a new dataset of automotive radar captures, and we validate the effectiveness of our approach on scene reconstruction and novel view synthesis on in-the-wild scenes.
\end{itemize}

\section{Related Work}
\label{sec:related}
\paragraph{Radar as a Sensing Modality} has become indispensable in object recognition and scene understanding, especially in robotic, maritime \cite{MaritimRadar}, and autonomous driving domains \cite{CramNet, BiDirectionalLiDARRadar, Bijelic_2020_STF}. The capability of mm-wave radiation in weather penetration \cite{CramNet, Bijelic_2020_STF} sets it apart from optical sensors. Beyond this, radar contributes significantly to constructing detailed environmental models that can predict depth \cite{DengxinRadarDepth}, semantics \cite{MultiViewSemanticSeg, Zhang_2023_CVPR}, scene flow \cite{RadarSceneFlow}, object presence \cite{CramNet, Kim_2023_ICCV, LiDARRadarFusionKitani, Bijelic_2020_STF} and non-line-of-sight imaging \cite{scheiner2019seeing}. When fused with camera \cite{DengxinRadarDepth, RadarSceneFlow} or LiDAR data \cite{LiDARRadarFusionKitani, CramNet}, the inherent sparsity of radar data can be mitigated. Other approaches for MIMO radar leverage beam steering techniques and neural networks to super-resolve measurements. \cite{CoIRRadarImaging, RadarResolution,FogMimoRadar}.
The progress in this field owes much to datasets \cite{RadarDatset2022MultiTaskValeoRadial, AutomotiveRadarDatasetAstyx,nuscenes}, which have provided invaluable data and benchmarks. We introduce a new dataset curated for radar neural rendering to further advance the state-of-the-art.

\paragraph{Radar Maps} provide a layer of information, in addition to LiDAR and camera reconstructions, allowing for improved understanding of the free space around an autonomous vehicle \cite{RadarMapping, SLAMFMCWLiDAR}. To this end, existing methods accumulate radar measurements and interpret them to classify if each individual grid cell is occupied, thereby enhancing scene reasoning capabilities \cite{RadarBasedMapping, RadarIndoor1}. This detailed grid mapping, combined with advanced radar perception, assists in translating raw radar data into interpretable occupancy information. Existing methods differentiate between static and dynamic entities and apply techniques like Doppler processing for object categorization based on relative velocity \cite{RadarDopplerAnalysis}. With challenges such as multipath interference, beam divergence, and clutter from unwanted reflections, signal post-processing techniques like adaptive filtering \cite{RadarGridmapRepresentationBeamDivergenceFilter} and polarimetric backscatter analysis \cite{RadarIndorMapping215Ghz} become crucial.
Our proposed representation substantially expands the convex hull of existing methods, and allows us to model object geometry due to the separation of radar cross-section into occupancy and reflectance.

\paragraph{Neural Rendering}
A large body of work uses single-sensor measurements to create detailed scene representations primarily from camera and LiDAR data \cite{zhang2023nerflidar,tao2023lidarnerf,huang2023nfl}. The use of radar data remains unexplored in this domain. Recognizing this gap, we propose the first neural rendering approach to generate geometric representations from radar data, paving the way for navigation and control applications using this representation in autonomous robotics and beyond.

Learned representations have driven recent advancements in novel-view generation \cite{mildenhall2020nerf,barron2021mipnerf,chen2022tensorf,muller2022instantNGP} and depth estimation \cite{tosi2023nerf4depth,tosi2023nerf}. Central to these advancements are neural radiance field methods \cite{mildenhall2020nerf,barron2021mipnerf,chen2022tensorf,muller2022instantNGP}, which model scenes as continuous volumetric fields of radiance. These techniques utilize volumetric rendering as a forward model, facilitating smooth interpolation between sensor poses. Various scene representations have been investigated for this task: coordinate-based networks \cite{mildenhall2020nerf,barron2021mipnerf,barron2022mipnerf360,zhang2021nerfactor}, 3D voxel-grids \cite{fridovich2022plenoxels,yu2021plenoctrees,chen2022tensorf}, and hybrid models \cite{muller2022instantNGP,barron2023zipnerf,tancik2023nerfstudio}. To enhance efficiency during both training and testing, approaches like \cite{yu2021plenoctrees,barron2023zipnerf,chen2022tensorf,muller2022instantNGP} have been developed.
Recent works have extended these techniques to capture vast outdoor scenes \cite{zhang2020nerf++,barron2022mipnerf360}. Representing extensive terrains, especially when captured by moving vehicles, presents unique challenges. This is due to the restricted and aligned views stemming from a single trajectory \cite{tancik2022blocknerf,turki2023suds,rematas2022urf,guo2023streetsurf,wang2023fegr,kundu2022panoptic,yang2023unisim,liu2023waabiICCV23,ost2022pointlightfields}. To overcome these challenges, researchers have incorporated supplemental supervisory cues, such as sparse LiDAR scans \cite{ost2022pointlightfields, turki2023suds, rematas2022urf,guo2023streetsurf}, estimated depths \cite{deng2022depthNeRF,roessle2022densedepthNeRF,guo2023streetsurf}, optical flow data \cite{turki2023suds,meuleman2023localrf}, and semantic segmentation \cite{kundu2022panoptic, wang2023fegr,turki2023suds}.

Recent work aims to learn scene representations from LiDAR data, which presents challenges due to its sparsity — often at two orders of magnitude less dense than camera data. These methods incorporate precise LiDAR point registration \cite{huang2023nfl}, or apply a two-tiered approach, using weak semantic supervision to filter out points below a registration threshold \cite{zhang2023nerflidar}. Others directly predict point drop likelihoods \cite{tao2023lidarnerf}.

Beyond optical sensors, non-optical sensors like imaging sonar \cite{SonarCMU, SonarASU} have leveraged neural representations and physics-based rendering techniques to achieve state of the art 3D reconstructions of single objects in isolation. Applying neural radiance methods to radar data arguably poses an even bigger challenge due to its lower density and long range that enable it to capture large outdoor scenes. We harness raw radar waveforms, which offer improved estimates of empty space.

\label{sec:method}
\section{Radar Fields}
\begin{figure*}[ht!]
\vspace{-10pt}
    \centering
    \includegraphics[width=0.92\linewidth]{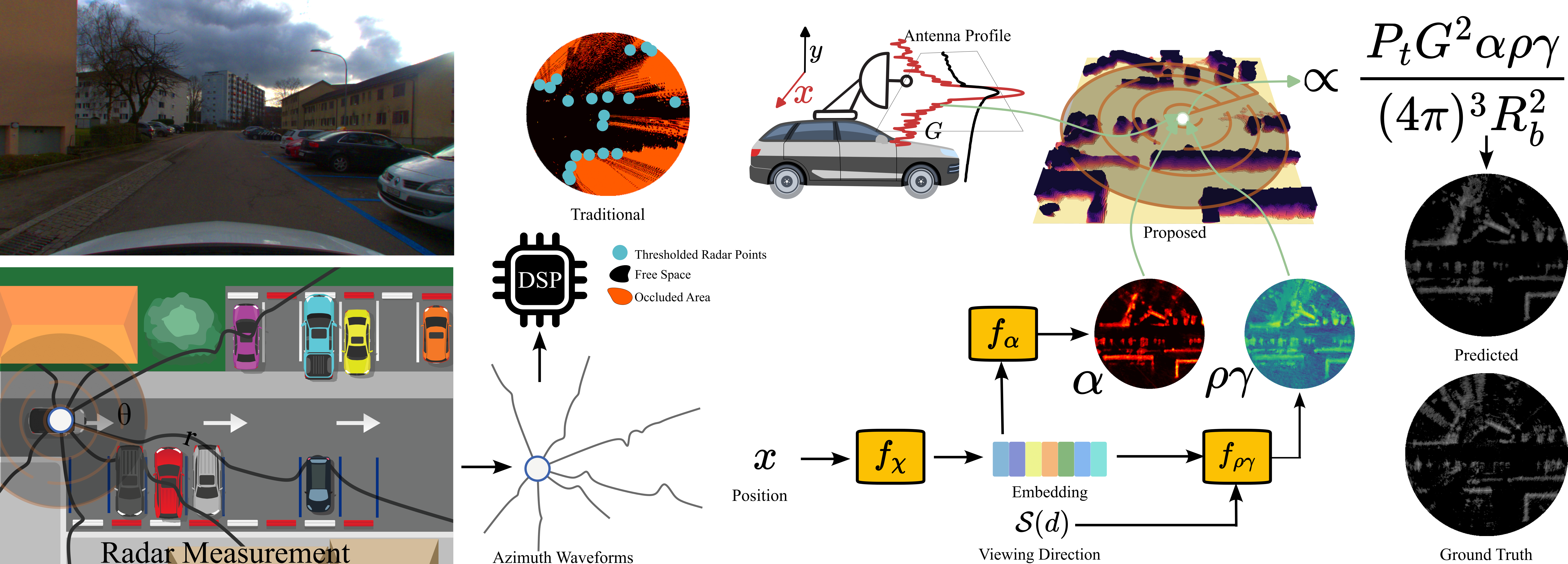}
    \vspace{-1eM}
    \caption{Radar Fields recovers a 3D scene from raw FFTs of 2D bird’s eye view radar scans. Each radar frame captures information centered around a 2D circular disk (bottom-left), from which a volume-rendering-free representation of the scene is learned (right). Following the antenna gain profile, points along super-sampled azimuthal rays are converted by \(f_\chi\) into an embedding. This embedding is subsequently processed by \(f_\alpha\) and \(f_{\rho\gamma}\), which decompose the signal intensity into occupancy \(\alpha\) and reflectance \(\rho\gamma\). To reconstruct FFT measurements, we integrate the super-sampled returns from both representations for each range-azimuth point, and apply our forward model (top-right).}
    \label{fig:overview}
    \vspace{-6pt}
\end{figure*}

We introduce Radar Fields as a scene representation capable of recovering dense geometry and synthesizing radar signals at unseen views by fitting to a \emph{single trajectory} of raw radar measurements. The method is illustrated in Fig.~\ref{fig:overview}. It hinges on a physics-based forward model that allows us to disentangle occupancy and reflectance (Sec.~\ref{sec:radar-formation}). Tailored to this physical model, we introduce a novel implicit radar field representation (Sec.~\ref{sec:radar-fields-rep}) and a physics-based importance sampling schema (Sec.~\ref{sec:radar-fields-sampling}). We fit the model by reconstructing raw radar signals in frequency space (Sec.~\ref{sec:radar-fields-loss}).

\subsection{Radar Signal Formation Model}
\label{sec:radar-formation}
Radar systems emit electromagnetic waves and analyze their reflections from objects to derive distance and velocity information. Frequency-modulated continuous-wave (FMCW) radar is distinct in that it emits a continuous radio waveform with a frequency that varies over time, often with a sawtooth-modulated `chirp' pattern \cite{FMCWTextbook}. Assuming sawtooth modulation, the frequency of the transmitted signal, $f_\tau$ changes linearly over time $t$ as
\begin{equation}
    f_\tau(t) = \omega + \theta(t), \;\;\; \text{with} \;\;\; \theta(t) = 2\Delta f \cdot \text{mod}\left(\frac{t}{T_s}, 1\right),
\end{equation}
where $\omega$ is the constant carrier frequency and $\theta(t)$ is a periodic sawtooth function with period $T_s$ and half-amplitude $\Delta f$.

When the chirp reflects off an object and returns to the sensor, the time delay introduced by the distance traveled results in a phase offset and therefore a frequency difference between the transmitted waveform $f_\tau$ and the received waveform $f_r$. For a single object,
\begin{equation}
    f_r(t) = \omega + \theta(t - t_d), \;\;\; \text{with} \;\;\; t_d = \frac{2R}{c},
\end{equation}
where $t_d$ is the two-way time delay for light reflecting off the object at range $R$ and returning to the radar, assuming that the target is stationary and the Doppler frequency shift is zero.

Both signals are then processed through a mixer and low-pass filter to compute their instantaneous frequency and phase differences. These computed differences can be treated as a new signal, called the intermediate frequency (IF) waveform, whose frequency $f_{IF}$ and phase are equal to the computed differences, respectively
\begin{equation}
    f_{IF}(t) = f_\tau(t) - f_r(t) = \theta(t) - \theta(t - t_d).
\end{equation}
In the case of a single target, the IF waveform is a sinusoid with frequency $f_{IF}$. But In practice, multiple objects are usually detected, meaning that the IF signal would be a sum of many different frequency sinusoids, each of which corresponds to the range and $f_r(t)$ of a different target \cite{FMCWTextbook}. Subsequently, a Fast Fourier Transform (FFT) yields the observable targets at any distance $R_b$,
\begin{equation}
    P_r(b) = \sum_{n=0}^{N_b - 1} IF(n) \cdot e^{\frac{-i2\pi bn}{N_b}}, \;\;\; \text{with} \;\;\;
    b = f_{IF_b},
\end{equation}
where $IF$ is the IF waveform and $N_b$ is the total number of frequency bins, such that each bin $b$ corresponds to a different tone in the IF frequency $f_{IF_b}$ and is correlated with a different range via
\begin{equation}
    R_b = f_{IF_b} \frac{cT_s}{4\Delta f}.
\label{eq:range_correlation}
\end{equation}

We aim to fit the raw radar FFT signal described above. As such, our method needs to predict the detected power $P_r(b)$ at every bin $b$, thereby reconstructing a frequency-space waveform for every azimuth-resolved beam. To this end, we rely on the known physics of FMCW radar to formulate a signal formation model.
Due to the correlation between range and IF frequency in Eq. \ref{eq:range_correlation}, the detected power in Fourier space at a given frequency bin $b$, $P_r(b)$, can be modeled as the returned power detected by the sensor at corresponding range $R_b$.
Therefore, we can assume that for an FFT bin $b$, whose frequency corresponds to a distributed target at range $R_b$ from the sensor, the received power $P_r(b)$ can be calculated as,
\begin{equation}
    P_r(b) = \frac{P_t \cdot G^2 \cdot \sigma }{(4\pi)^3 R_b^2},
    \label{eq:forwardEq}
\end{equation}
where $P_t$ is the transmitted power, $G$ is the antenna gain and, $\sigma$ is the total radar cross-section of the distributed target \cite{RichardsTextbook, skolnik2001introduction}. Note that every term in this equation is known except for $\sigma$. The radar cross-section considers object shape, material, and reflective properties, describing how detectable it is by a radar.

However, predicting $\sigma$ directly is insufficient for geometry reconstruction.
Instead, we aim to disentangle the geometry-dependent component of $\sigma$ from its specular and material components. The radar cross section $\sigma$ can be further decomposed into three terms, as 
$\sigma = \alpha \cdot \rho \cdot \gamma $,
with $\alpha$ being the object size projected onto the cross-section of the radar beam, and with $\rho$ being the reflectivity and $\gamma$ being the directivity of the object. Here, $\alpha$ depends solely on object geometry, while $\rho$ and $\gamma$ also depend on how metallic an object is, as well as the incident angle of the radar beam. This defines an interpretable forward model as the foundation of this method, where separate neural fields can be learned for each term.

\begin{figure*}[t]
\vspace{-2pt}
    \centering 
    \includegraphics[width=\linewidth]{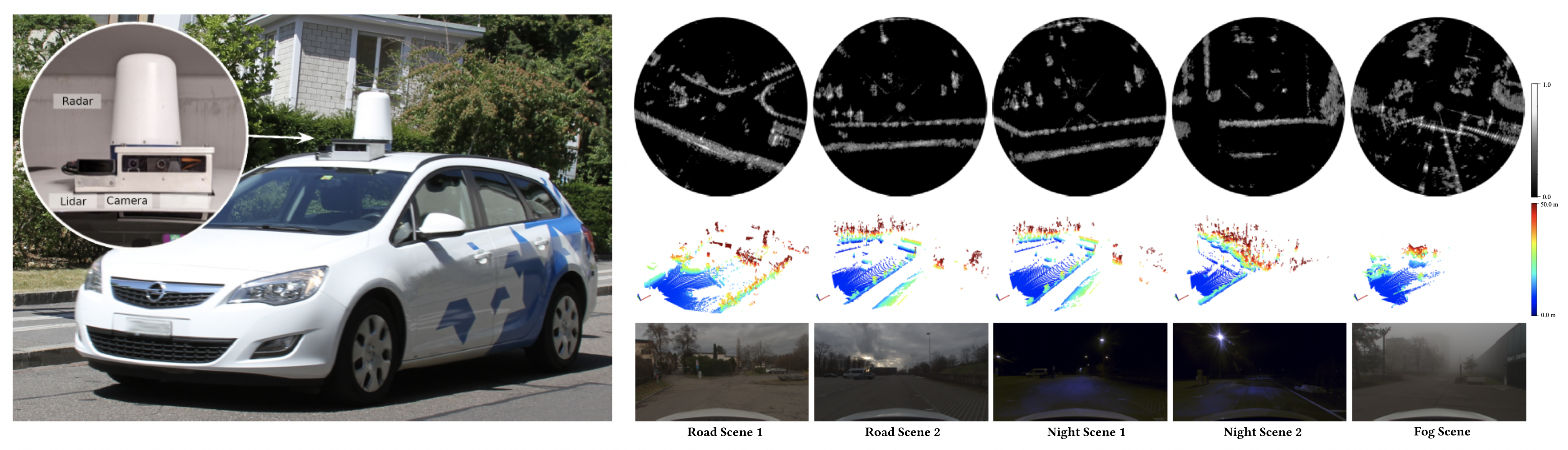}\vspace{-9pt}
    \caption{Multi-modal Dataset for Training and Validation. The waterproof sensor rig (left) which is used to collect our dataset and 5 exemplary scenes (right) with (1\textsuperscript{st} row) 40-meter-radius BEV radar returns with the car in the center and pointing to the right, and (2\textsuperscript{nd} row) point clouds from our forward-facing LiDAR, color-coded by height, with the car in the bottom-left and (3rd row) images from our forward-facing RGB camera.}
    \label{fig:sensor_rig}
    \vspace{-10pt}
\end{figure*}

\subsection{Implicit Neural Field Representation}
\label{sec:radar-fields-rep}
We learn an implicit neural model to render our two-dimensional radar measurements $P_r \in \mathbb{R}^{N_\phi \times N_b}$ for $N_b$ range bins and $N_\phi$ azimuth angles. The radar cross-section $\sigma \in \mathbb{R}$ is predicted for each measurement grid cell. $\sigma$ is decomposed into projected cross-sectional area $\alpha \in \mathbb{R}$, which represents scene occupancy, and  the product of reflectivity and directivity, $\rho\gamma \in \mathbb{R}$, which represents scene reflectance. We fit our model to a sequence of raw radar captures, and corresponding poses from an on-board Global Navigation Satellite System (GNSS), as in Fig. \ref{fig:overview}.

We reconstruct ground truth FFT data using two neural fields, $f_\alpha$ and $f_{\rho\gamma}$, which represent scene occupancy and reflectance, respectively. Both fields are conditioned on an embedding $\chi\ \in \mathbb{R}^d$ from the neural field $f_\chi$.
Given a point in space $\mathbf{x} \in \mathbb{R}^3$ and view direction $\mathbf{d} \in \mathbb{R}^3$, the total scene decomposition can be written as
\begin{align*}
f_{\chi}:\{\mathcal{H}(\mathbf{x})\} \longrightarrow & \{\chi\} \text { Feature Embedding }\\ \nonumber
f_{\alpha}:\{\chi\} \longrightarrow & \{\alpha\}  \text { Occupancy } \\ \nonumber
f_{\rho\gamma}:\{\mathcal{S}(\mathbf{d}), \chi\} \longrightarrow &\{\rho\gamma\} \text { Reflectance }\nonumber
\end{align*}
where $\mathcal{H}$ and $\mathcal{S}$ are multi-resolution hash encodings as in \citet{HashGrid}, and spherical harmonic positional encoding as in \citet{SphericalHarmonics}, respectively.

The embedding field $f_{\chi}:\{\mathcal{H}(\mathbf{x})\} \longrightarrow\{\chi\}$ predicts geometric and material features $\chi$ from hashgrid-encoded position.
These latent scene features are shared by both components of our scene representation.
We predict occupancy with field $f_{\alpha}:\{\chi\} \longrightarrow\{\alpha\}$ as projected cross-sectional area $\alpha$ onto the radar beam.
We predict reflectance with field $f_{\rho\gamma}:\{\mathcal{S}(\mathbf{d}), \chi\} \rightarrow\{\rho\gamma\}$, which is also conditioned on view direction $\mathbf{d}$ encoded with spherical harmonics $\mathcal{S}$.
This field directly predicts the product of surface reflectivity and directivity, $\rho\cdot \gamma$, as there is no clear way to disambiguate these two values and supervise them independently. 

One advantage of this representation is its versatility, as $\alpha$ can be used alone to reconstruct basic scene geometry, but can also be multiplied with $\rho \cdot \gamma$ at unseen locations or novel views to \emph{synthesize returned waveforms directly}.
It is worthwhile to highlight that this model does \emph{not require volume rendering}. The ground truth radar signal is a raw time-resolved waveform and the ray samples themselves are the predictions.

\subsection{Physics-Based Importance Sampling}
\label{sec:radar-fields-sampling}

Radar beams diverge as they travel through a scene.
This divergence is not negligible in FMCW radar sensing, given both the long range capabilities of radio detection and the large elevation opening angles in automotive settings.
Note that most low-cost FMCW radars do not have any elevation resolution.
However, these sensors still capture information in three dimensions, as they integrate signals across elevation and azimuth for each beam.
We model these properties by super-sampling additional rays within the elliptical cone defined by our radar azimuth and elevation opening angles, allowing us to model our ground truth range-azimuth measurements in 3D.

FMCW radar systems do not use isotropically-radiating antennas that would distribute and receive power evenly across the opening angles of the beam.
Instead, they rely on antennas with a bias towards a specific direction.
These exhibit the highest signal gain in the direction of the beam center while dropping off steeply at increasingly divergent angles off-boresight.
This can be represented as a radiation pattern, which measures signal gain as a function of angular offset relative to the center of a beam, and affects returned power as in Eq.~\ref{eq:forwardEq}. We consider two radiation patterns, $\mathcal{A}(a)$ and $\mathcal{E}(e)$, which map azimuth and elevation angular offsets $a$ and $e$ to signal gain, respectively.

To model these sensor properties, we super-sample additional rays, distributed uniformly within the elliptical cone of beam divergence around each beam center.
For each of the $N$ beam centers randomly sampled at any given training step, we sample an additional $S-1$ rays to create a set $\mathbb{S} = \left\{s_1, s_2, ..., s_S \right\}$ such that $|\mathbb{S}| = S$. Each ray super-sample consists of angular offsets $a_i$ and $e_i$ from the beam center, such that $s_i = \left\{a_i,e_i\right\} \forall s_i \in \mathbb{S}$. The super-samples $s_i$ are drawn from a pair of uniform distributions, such that $ a_i \sim \mathcal{U}(-\mathrm{A}, \mathrm{A}) \text{ and } e_i \sim \mathcal{U}(-\mathrm{E}, \mathrm{E})$, where $\mathcal{U}$ is a uniform distribution over all of the possible angular offsets from the beam center with azimuthal field of view $2\mathrm{A}$ and elevation field of view $2\mathrm{E}$. 
We query our model with $S$ uniformly-distributed rays per beam and
 predict a separate radar cross-section $\hat{\sigma_i} = \hat{\alpha_{i}} \cdot \hat{\rho\gamma_i}$, for each super-sampled ray $s_i \in \mathbb{S}$ and average them proportional to our sensor radiation profiles $\mathcal{A}$ and $\mathcal{E}$ to compute the final predicted radar cross-section $\hat{\sigma}$ at each range bin,
\begin{equation}
    \hat{\sigma} = \frac{\sum_{i=1}^{S} \sigma_i \cdot \mathcal{A}(s_i[0]) \cdot \mathcal{E}(s_i[1])}{\sum_{i=1}^{S} \mathcal{A}(s_i[0]) \cdot \mathcal{E}(s_i[1])},
\end{equation}
from which we compute the predicted power return.
This way, samples that lie further from each beam center contribute less to the received signal intensity, mirroring the physics of the radar measurement process and establishing a link between our learned 3D scene representation and our 2D ground truth data.

\subsection{Training}
\label{sec:radar-fields-loss}

We train our model to reconstruct a scene from a sequence of radar frames applying the total loss $\mathcal{L}$.
The sole supervision signal is the raw radar waveform itself. Predicted occupancy is also regularized to enforce its geometric interpretation.

The total loss $\mathcal{L}$ is a weighted sum of the reconstructed FFT loss $\mathcal{L}_W$, and the regularization terms $\mathcal{L}_R$ and $\mathcal{L}_P$ for the occupancy, leading to,
\begin{align}
    \mathcal{L} =& \; \eta_{_{W}}\mathcal{L}_W + \eta_{_{R}}\mathcal{L}_R + \eta_{_{P}}\mathcal{L}_P\\
    \mathcal{L}_W =& \; \frac{1}{N_\phi N_b}\sum_{\phi,b} \left\| \frac{P_t \cdot G^2 \cdot (\hat{\alpha}_{\phi,b}\cdot \hat{\rho\gamma}_{\phi,b}) }{(4\pi)^3 R_b^4} - P_{r_{\phi,b}}\right\|\\
    \mathcal{L}_R =& \; \frac{1}{BatchSize}\sum_{\phi,b} \mathcal{O}(P_r)_{\phi,b} \left(log(\mathcal{O}(P_r)_{\phi,b}) - log(\hat{\alpha}_{\phi,b}) \right )\\
    \mathcal{L}_P =& \; \text{std}\left(\hat{\alpha}\rvert_{\mathcal{O}(P_r) > 0.5}\right) + \text{std}\left(\hat{\alpha}\rvert_{\mathcal{O}(P_r) < 0.5}\right)
\end{align}
where the weights are $\eta_{_{W}}, \eta_{_{R}}$ and $\eta_{_{p}}$. 
In detail, $L_W$ assesses the quality of the reconstructed FFT signal.
$\mathcal{L}_R$ ensures that the learned scene geometry aligns with the occupancy derived from the raw signal. Occupancy from a 2D raw radar signal can be estimated by applying an occupancy estimator $\mathcal{O}(P_r) \in \mathbb{R}^{N_\phi \times N_b}$. Our occupancy probability estimation algorithm $\mathcal{O}(P_r)$ follows previous work, like \cite{grid_mapping}, scanning over ground truth bins for each frame and estimating occupancy likelihood per-bin with a simple Bayesian update rule and occlusion model. More details are in the Supplementary Material.
$\mathcal{L}_P$ enforces a bimodal distribution of occupancy probabilities, ensuring that empty space is correctly modeled and removing floater artifacts.

To prevent the model from over-fitting to noisy data and getting trapped in local minima, we employ a coarse-to-fine optimization procedure, as in \cite{CoarseToFine}. We mask our multi-resolution hashgrid encodings $\mathcal{H}(\mathbf{x})$ as 
\begin{equation}
\mathcal{H}(\mathbf{x}) = \begin{cases}
\mathcal{H}(\mathbf{x}) & \text{ if } \frac{i}{|\gamma|} < 0.4+0.6\left(\text{sin$\left(\frac{\text{epoch}}{\text{max. epoch}}\right)$}\right)\\
0 & \text{ otherwise}
\end{cases},
\end{equation}
gradually providing the model with higher-resolution grid features over the course of training. This way, reconstruction noise cannot accumulate in higher frequency features during early training.

\section{Dataset}
\label{sec:dataset}
We train and validate our method on a novel multi-modal dataset. The recorded modalities include raw radar, LiDAR, RGB camera, and GNSS, with their sensor specifications described in Tab. 1 of the Supplementary Material. Our FMCW radar, which is a custom Navtech CIR-DEV, recording 360\textdegree {} around the vehicle and providing a long range of 330 m at a range resolution of ca.\ 4 cm. The camera, a TRI023S-C with 2 MPix, and the LiDAR both span a frontal field of view (FOV). 
We incorporate a state-of-the-art microelectromechanical-system (MEMS) LiDAR (RS-LiDAR-M1) with a horizontal FOV of 120\textdegree.

To ensure accurate sensor supervision, we performed both geometric calibration and synchronization of our sensors.
We calibrate the RGB camera intrinsics with conventional checkerboard calibration~\cite{opencv_library}, while the intrinsic calibration for both LiDAR and radar is provided by the vendor.  
Our extrinsic calibration for LiDAR-camera uses mutual information maximization~\cite{pandey2015automatic}. For radar-LiDAR calibration, we follow~\cite{burnett2022boreas} and estimate the rotation via correlative scan matching using the Fourier Mellin transform~\cite{checchin2010radar}, with the translation directly measured. GNSS-LiDAR calibration uses u-center~\cite{u:center} and LiDAR point cloud consistency optimization.
We synchronize all internal clocks and record each sensor independently.

We completed extensive in-the-wild driving sessions in Switzerland, capturing multiple hours of footage.
We selected 15 sequences, emphasizing scene diversity and ensuring a high quality GNSS signal. Each sequence has a duration of 10-23 seconds, containing 40-90 radar frames. These authentically represent real-world driving scenarios, with the vehicle moving at speeds from 5 to 30 km/h.

The dataset includes diverse conditions, including day, night, and fog, and types of scenes, including urban streets and parking lots, with example scenes visualized in Fig.~\ref{fig:sensor_rig}. This diversity allows us to investigate the impact of varying scenes and environmental conditions.
We include recordings from both day and night in the same scene, enabling detailed examinations of ambient light effects.
We investigate sensor disparities between radar and LiDAR by including adverse weather scenes recorded in strong fog.
We anticipate the radar maintaining performance in these adverse conditions, while the LiDAR and camera are expected to be more affected. 

\label{sec:results}
\section{Assessment}

In this section, we assess the proposed method with the novel multimodal dataset described in the previous section. Specifically, we investigate scene reconstruction across day, night, and fog scenes, for bird's eye view (BEV) occupancy reconstruction and novel view synthesis in 2D, and with geometry reconstruction in 3D. Specifically, we validate that the method is capable of recovering 3D scene representations from conventional 2D radar scans by incorporating the angular-dependent antenna response in the forward model.

\subsection{Experimental Setup}
\label{subsec:exp-setup}

The collected outdoor scenes in our dataset fall into three main categories: parking lots featuring a high number of dielectric surfaces on parked vehicles that appear in radar returns, urban, where scene composition is complex, including vehicles, and  adverse weather scenes, including rain and fog. We withhold 20 \% of all frames, consecutively, in a single observation gap to form a train-test split.

\begin{table*}[ht!]
\caption{Quantitative Assessment. We measure geometry and reconstruction accuracy of our method compared to a radar grid mapping baseline \cite{grid_mapping} and LiDAR-NeRF~\cite{tao2023lidarnerf}. We also validate the effectiveness of importance sampling \& regularization as an ablation. The proposed method compares favorably in CD and RCD on the scene reconstruction task (see text for metrics), and in terms of RMSE and PSNR for novel view synthesis. These metrics confirm the findings from Fig \ref{fig:results}. We note our method outperforms LiDAR-NeRF in adverse weather conditions as visibly proven in Fig.~\ref{fig:fog-fig}.}
\vspace{-5pt}
\resizebox{0.9999\linewidth}{!}{
\begin{tabular}{ccccccccccccccccccccccccc}
\toprule
& \multicolumn{4}{c}{\textbf{Scene 1}}                                
& \multicolumn{4}{|c}{\textbf{Scene 2}}                                
& \multicolumn{4}{|c}{\textbf{Scene 3}}                               
& \multicolumn{4}{|c}{\textbf{Scene 4}}                                
& \multicolumn{4}{|c}{\textbf{Scene 5}}                                
& \multicolumn{4}{|c}{\textbf{Scene Fog}}                               \\
   
& \multicolumn{1}{c}{\textbf{CD $\downarrow$}} & \multicolumn{1}{c}{\textbf{RCD $\downarrow$}} & \multicolumn{1}{c}{\textbf{RMSE$\downarrow$}} & \multicolumn{1}{c}{\textbf{PSNR $\uparrow$}}
& \multicolumn{1}{|c}{\textbf{CD $\downarrow$}} & \multicolumn{1}{c}{\textbf{RCD $\downarrow$}} & \multicolumn{1}{c}{\textbf{RMSE$\downarrow$}} & \multicolumn{1}{c}{\textbf{PSNR $\uparrow$}}
& \multicolumn{1}{|c}{\textbf{CD $\downarrow$}} & \multicolumn{1}{c}{\textbf{RCD $\downarrow$}} & \multicolumn{1}{c}{\textbf{RMSE$\downarrow$}} & \multicolumn{1}{c}{\textbf{PSNR $\uparrow$}}
& \multicolumn{1}{|c}{\textbf{CD $\downarrow$}} & \multicolumn{1}{c}{\textbf{RCD $\downarrow$}} & \multicolumn{1}{c}{\textbf{RMSE$\downarrow$}} & \multicolumn{1}{c}{\textbf{PSNR $\uparrow$}}
& \multicolumn{1}{|c}{\textbf{CD $\downarrow$}} & \multicolumn{1}{c}{\textbf{RCD $\downarrow$}} & \multicolumn{1}{c}{\textbf{RMSE$\downarrow$}} & \multicolumn{1}{c}{\textbf{PSNR $\uparrow$}}
& \multicolumn{1}{|c}{\textbf{CD $\downarrow$}} & \multicolumn{1}{c}{\textbf{RCD $\downarrow$}} & \multicolumn{1}{c}{\textbf{RMSE$\downarrow$}} & \multicolumn{1}{c}{\textbf{PSNR $\uparrow$}} \\

& \multicolumn{1}{c}{\textbf{$\left[m\right]$}} & \multicolumn{1}{c}{-} & \multicolumn{1}{c}{-} & \textbf{$\left[dB\right]$}
& \multicolumn{1}{|c}{\textbf{$\left[m\right]$}} & \multicolumn{1}{c}{-} & \multicolumn{1}{c}{-} & \textbf{$\left[dB\right]$}
& \multicolumn{1}{|c}{\textbf{$\left[m\right]$}} & \multicolumn{1}{c}{-}  & \multicolumn{1}{c}{-} & \textbf{$\left[dB\right]$}
& \multicolumn{1}{|c}{\textbf{$\left[m\right]$}} & \multicolumn{1}{c}{-}  & \multicolumn{1}{c}{-} & \textbf{$\left[dB\right]$}
& \multicolumn{1}{|c}{\textbf{$\left[m\right]$}} & \multicolumn{1}{c}{-}  & \multicolumn{1}{c}{-} & \textbf{$\left[dB\right]$}
& \multicolumn{1}{|c}{\textbf{$\left[m\right]$}} & \multicolumn{1}{c}{-}  & \multicolumn{1}{c}{-} & \textbf{$\left[dB\right]$}\\

\midrule
Proposed       
& \multicolumn{1}{c}{\textbf{0.163}}     & \multicolumn{1}{c}{0.013}     & \multicolumn{1}{c}{\textbf{0.185}}     &  \textbf{20.660}  
& \multicolumn{1}{c}{\textbf{0.296}}     & \multicolumn{1}{c}{\textbf{0.007}}     & \multicolumn{1}{c}{\textbf{0.190}}     &   \textbf{20.431}   
& \multicolumn{1}{c}{0.166}     & \multicolumn{1}{c}{\textbf{0.022}}     & \multicolumn{1}{c}{\textbf{0.169}}     &   \textbf{21.461}    
& \multicolumn{1}{c}{\textbf{0.227}}     & \multicolumn{1}{c}{\textbf{0.025}}     & \multicolumn{1}{c}{\textbf{0.198}}     &   \textbf{20.108}
& \multicolumn{1}{c}{\textbf{0.193}}     & \multicolumn{1}{c}{\textbf{0.018}}     & \multicolumn{1}{c}{\textbf{0.184}}     &     \textbf{20.738}
& \multicolumn{1}{c}{0.382}     & \multicolumn{1}{c}{\textbf{0.015}}     & \multicolumn{1}{c}{\textbf{0.185}}     &     \textbf{20.738}\\
Ablation
& \multicolumn{1}{c}{0.260}     & \multicolumn{1}{c}{0.014}    & \multicolumn{1}{c}{0.231}    &     18.740
& \multicolumn{1}{c}{0.482}     & \multicolumn{1}{c}{0.012}     & \multicolumn{1}{c}{0.262}     &     17.647
& \multicolumn{1}{c}{0.243}     & \multicolumn{1}{c}{0.023}     & \multicolumn{1}{c}{0.199}     &     20.030
& \multicolumn{1}{c}{0.284}     & \multicolumn{1}{c}{0.026}     & \multicolumn{1}{c}{0.278}     &     17.132
& \multicolumn{1}{c}{0.194}     & \multicolumn{1}{c}{0.018}     & \multicolumn{1}{c}{0.261}     &     17.691
& \multicolumn{1}{c}{\textbf{0.013}}     & \multicolumn{1}{c}{0.356}     & \multicolumn{1}{c}{0.217}     &     19.32 \\
Grid Mapping ~\shortcite{grid_mapping} 
& \multicolumn{1}{c}{0.240}     & \multicolumn{1}{c}{0.019}     & \multicolumn{1}{c}{0.212}    &    19.492
& \multicolumn{1}{c}{1.222}     & \multicolumn{1}{c}{0.047}     & \multicolumn{1}{c}{0.228}     &     18.850
& \multicolumn{1}{c}{\textbf{0.164}}     & \multicolumn{1}{c}{0.027}     & \multicolumn{1}{c}{0.220}     &     19.171
& \multicolumn{1}{c}{0.241}     & \multicolumn{1}{c}{0.036}     & \multicolumn{1}{c}{0.206}     &     19.732
& \multicolumn{1}{c}{0.361}     & \multicolumn{1}{c}{0.033}     & \multicolumn{1}{c}{0.248}     &     18.148
& \multicolumn{1}{c}{0.614}     & \multicolumn{1}{c}{0.048}     & \multicolumn{1}{c}{0.255}     &     17.887\\
LiDAR-NeRF ~\shortcite{tao2023lidarnerf} 
& \multicolumn{1}{c}{0.189}     & \multicolumn{1}{c}{\textbf{0.012}}     & \multicolumn{1}{c}{-}    &    -
& \multicolumn{1}{c}{0.312}     & \multicolumn{1}{c}{0.008}     & \multicolumn{1}{c}{-}     &     -
& \multicolumn{1}{c}{0.187}     & \multicolumn{1}{c}{\textbf{0.022}}     & \multicolumn{1}{c}{-}     &     -
& \multicolumn{1}{c}{0.325}     & \multicolumn{1}{c}{0.093}     & \multicolumn{1}{c}{-}     &     -
& \multicolumn{1}{c}{0.399}     & \multicolumn{1}{c}{0.083}     & \multicolumn{1}{c}{-}     &     -
& \multicolumn{1}{c}{0.797}     & \multicolumn{1}{c}{0.087}     & \multicolumn{1}{c}{-}     &     -\\
\midrule
\bottomrule
\end{tabular}
}
\label{tab:quantitative_results_draft}
\vspace{-10pt}
\end{table*}

\paragraph{Evaluation Criteria}
\label{subsec:eval}
We evaluate our method both with quantitative metrics and qualitatively. We measure the PSNR and RSME of the reconstructed radar FFT signal compared to the withheld ground-truth radar FFT data. To assess the quality of BEV occupancy reconstructions, we also report the Chamfer Distance (CD) and Relative Chamfer Distance (RCD) between the 2D BEV point cloud estimated with our method and the one from the ground truth data. Note that estimating ground truth geometry from radar data alone is challenging due to the low angular sampling, entangled multipath effects, and sensor noise. Therefore, we additionally derive ground truth scene geometry from LiDAR, which provides geometry with a resolution that is an order of magnitude higher. We accumulate LiDAR data for the entire scene to create a dense ground truth, which is filtered and projected into the radar frame and combined with highly probable radar predictions.

\subsection{Scene Reconstruction}
\begin{figure}[t!]
\vspace{-3pt}
    \centering
    \includegraphics[width=\linewidth]{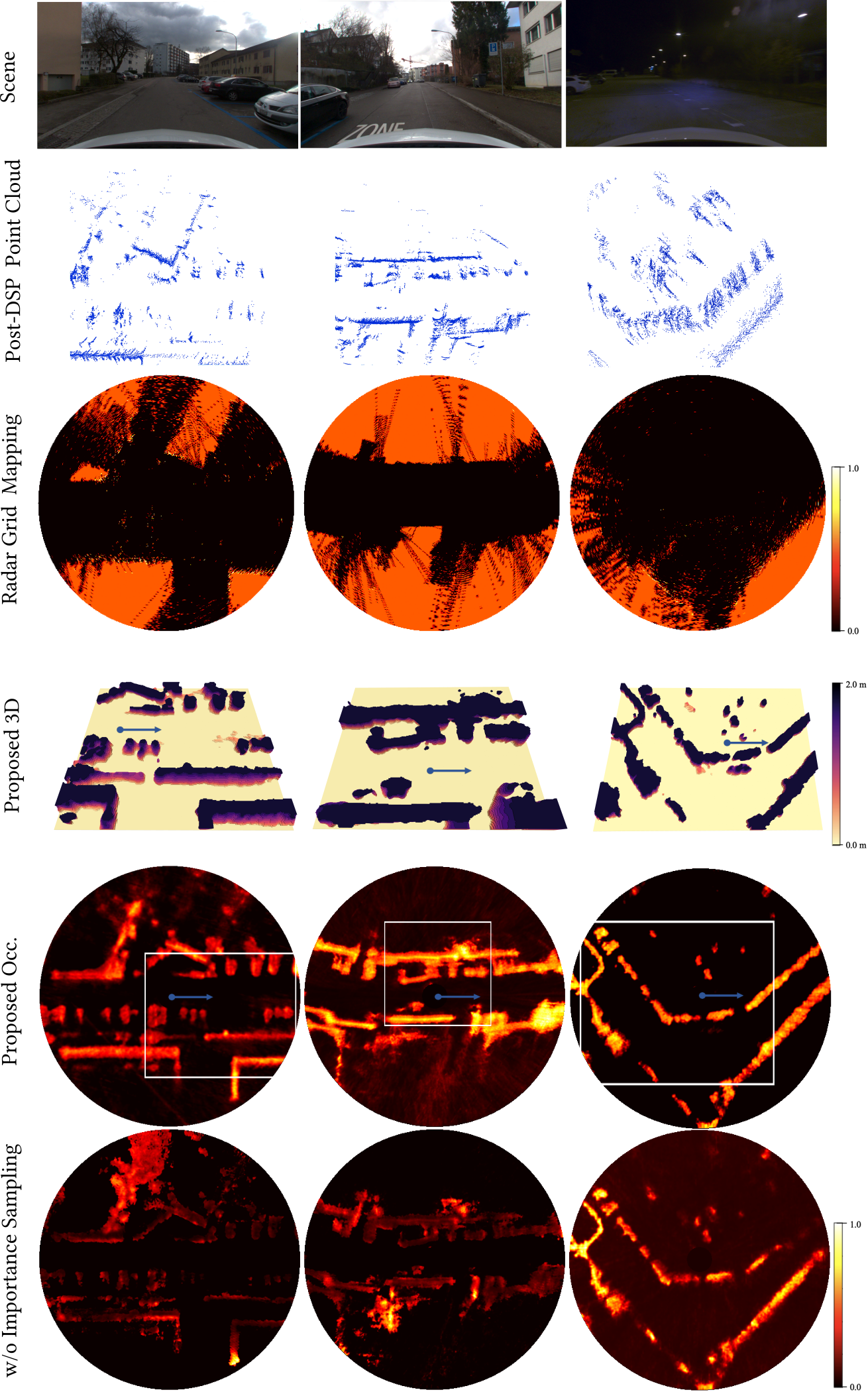}\vspace{-10pt}
    \caption{Radar Fields for Scene Reconstruction. Conventional post-processed radar point clouds (second row) are sparse, and, hence, conventional grid mapping methods \cite{grid_mapping} (third row) fail to recover accurate geometry. Radar Fields relies on raw frequency-space radar measurements and recovers high-quality BEV occupancy (fourth row), and even accurate 3D geometry (third row) from the same 2D radar scans. Without physics-based ray importance sampling (last row), the predicted occupancy becomes inconsistent.
    }
    \label{fig:results}
    \vspace{-5pt}
\end{figure}

We report qualitative scene reconstruction results in Fig.~\ref{fig:results}. We observe that 
accumulated radar point clouds from conventional DSP processing are too sparse to provide dense scene information, and therefore the occupancy grid-maps recovered from these inputs are unreliable and minimally informative, only tracing out a rough contour of large structures in the immediate vicinity of the sensor. Row three of Fig.~\ref{fig:results} shows that these types of methods completely fail to reconstruct any vehicles in the scene - which is problematic for safety-critical automotive settings - as the inconsistent multipath returns from these small objects average out instead of coinciding into a dense silhouette. These methods are largely unable to see behind occluding structures. Moreover, the 2D nature of processed radar point clouds confine occupancy to BEV, as there are no 3D cues available. Access to raw FFT data makes it possible for Radar Fields to reconstruct the scene, including all vehicles in test scenes. Crucially, our neural occupancy field is 3D. Our physics-based ray importance sampling, described in section \ref{sec:radar-fields-sampling}, models radio beam divergence to extract 3D information from a sensor with no elevation resolution. In these 3D reconstructions, it is again possible to distinguish vehicles and walls that were not visible in previous approaches. The last row in Fig.~\ref{fig:results} validates that the importance sampling and regularization terms are essential. Without them, our method struggles to disentangle occupancy from reflectance, as specular effects become more prevalent in the occupancy field.

In Table \ref{tab:quantitative_results_draft}, we use CD and RCD as metrics to evaluate BEV scene reconstruction. Confirming our findings from qualitative evaluation, our method consistently improves on grid mapping \cite{grid_mapping} with conventional radar post-processing. The metrics also validate the physics-based importance sampling and regularization.

\subsection{Adverse Weather}
\begin{figure}[t!]
\vspace{-3pt}
    \centering
    \includegraphics[width=\linewidth]{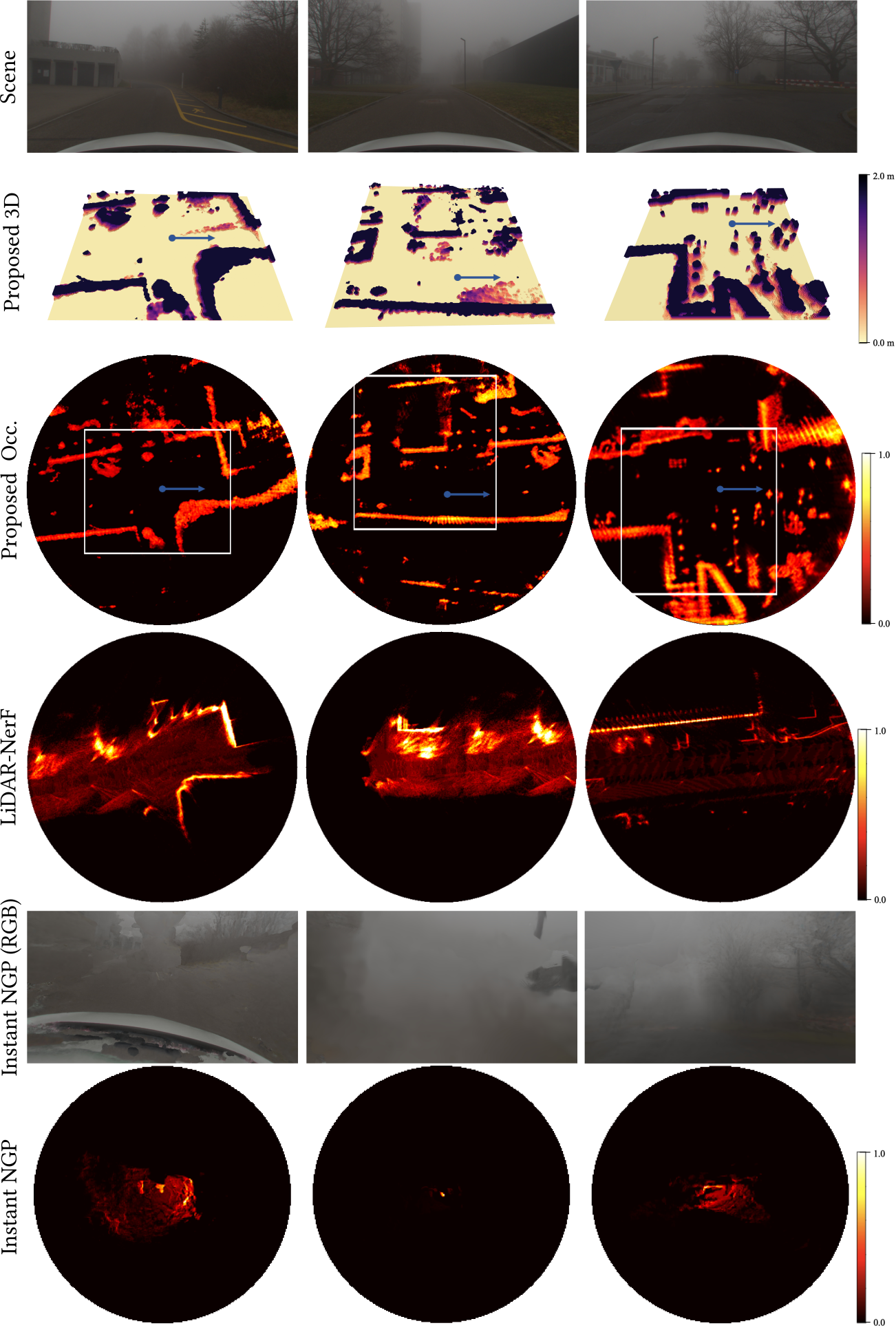}\vspace{-10pt}
    \caption{Radar Fields for adverse weather. Our method is robust to extreme weather and lighting conditions, including low-light and foggy scenes. RGB captures are reported in the top row. We compare our method to LiDAR-NeRF~\cite{tao2023lidarnerf} for LiDAR and Instant-NGP~\cite{muller2022instantNGP} for camera inputs as alternative modalities. LiDAR reconstructions struggle to capture accurate scene geometry due to backscatter. Multi-view reconstruction via RGB fails for this monocular foggy trajectory, also indicated by the synthesized RGB frames (second to last row). Radar Fields exhibits minimal degradation compared to good weather conditions.}
    \label{fig:fog-fig}
\end{figure}

Radar Fields retains reconstruction quality across weather and lighting conditions. Fig.~\ref{fig:fog-fig} reports reconstructions in fog, validating that the proposed method is able to reliably recover both 2D occupancy and 3D geometry in extreme conditions where other methods using LiDAR or camera input fail. We compare Radar Fields against LiDAR-NeRF~\cite{tao2023lidarnerf} using LiDAR input and Instant-NGP \cite{muller2022instantNGP} relying on RGB camera input, trained on the same day-time and night-time foggy scenes from our dataset. In these extreme conditions, LiDAR-NeRF performance degrades due to limited range information in dense fog, while Instant-NGP fails to reconstruct any meaningful occupancy due to the severe scattering in the scene. In contrast Radar Fields recovers crisp outlines of buildings, steel road barriers, and even vehicles remain visible. 
The last scene in Table \ref{tab:quantitative_results_draft} shows metrics for fog. Our method achieves lower CD and RCD than LiDAR-NeRF for scene reconstructions.

\subsection{Novel Radar View Synthesis}

\begin{figure*}[b!]
\vspace{-19pt}
    \centering
    \includegraphics[width=\linewidth]{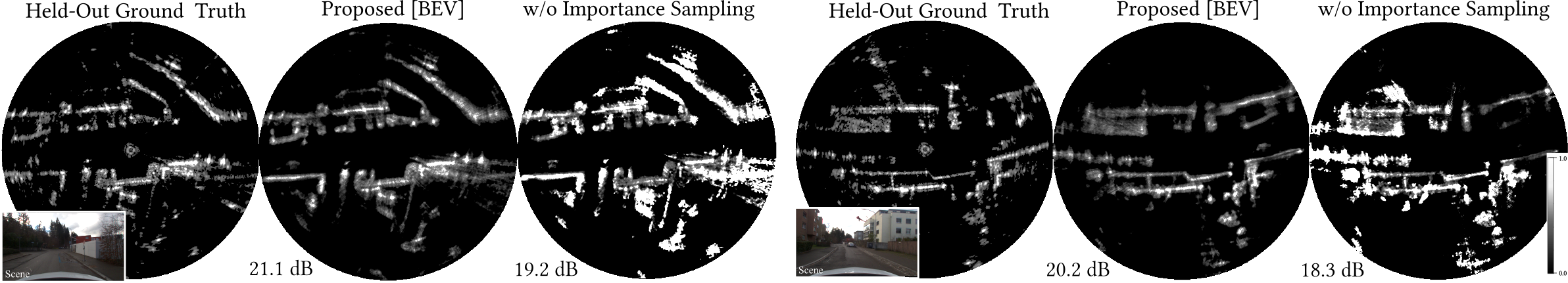}\vspace{-5pt}
    \caption{ Radar Fields for Novel Radar View Synthesis.
    Radar Fields is capable of synthesizing high-quality raw radar FFT measurements at novel viewpoints. Without our proposed super-sampling procedure, predicted measurements become noise prone and inaccurate in magnitude.
    }
    \label{fig:radar-NVS}
    \vspace{-5pt}
\end{figure*}

Radar Fields is grounded in reconstructing raw frequency waveforms, which is our only source of scene information. As such, it is capable of synthesizing raw radar returns at novel views. Fig.~\ref{fig:radar-NVS} confirms that these synthesized views can capture view-dependent reflective artifacts, like specular highlights and ray saturation.
We validate the method with RMSE and PSNR as quantitative metrics. Table \ref{tab:quantitative_results_draft} confirms also that the absence of importance sampling and regularization terms in the model leads to less accurate signal prediction for novel views and inconsistent convergence.


\section{Conclusion}
We introduce a neural rendering method for raw radar data. While a large body of work on neural scene representations investigates the reconstruction and generation of novel views from RGB and LiDAR point cloud data, neural rendering for radar measurements has been unexplored. Operating at millimeter wavelengths, radar sensors provide a signal complementary to optical imaging techniques -- radar signals penetrate fog and smoke with scattering particle sizes smaller than the wavelength. Unfortunately, the longer wavelength also inherently limits the angular resolution which existing neural rendering methods fundamentally rely on in multi-view consistent training. The proposed method tackles the resolution limitations of processed radar data and the computational cost of volume rendering by modeling scene parameters in frequency space. We supervise our models with raw radar waveform data in Fourier frequency space to recover relationships between detected power and distance from the sensor. We validate our method across diverse scenarios, especially in urban environments with dense vehicles and infrastructure, where mm-wavelength sensing is favorable. As a first step towards practical scene representations for radar, in the future, the proposed fusion methods could benefit from cross-modal input and supervision with LiDAR data, thereby bridging the resolution disparity, and optimizing geometry reconstruction despite the inherent angular resolution limitations of radar data.

\begin{acks}
This work was supported by the AI-SEE project with funding from the FFG, BMBF, and NRC-IRA. Eric Liang was supported by NSF GRFP (2146752). Felix Heide was supported by an Amazon Science Research Award, Packard Foundation Fellowship, Sloan Research Fellowship, Sony Young Faculty Award, the Project X Fund, and NSF CAREER (2047359). Authors from ETH Zurich were supported by the ETH Future Computing Laboratory (EFCL), financed by a donation from Huawei Technologies. The authors thank Julian Ost for fruitful discussions.
\end{acks}

\bibliographystyle{ACM-Reference-Format}
\bibliography{reference}

\end{document}